\documentclass[runningheads]{llncs}
\usepackage[T1]{fontenc}
\usepackage{graphicx}
\usepackage{amsmath,amssymb,amsfonts}
\usepackage[T1]{fontenc}
\usepackage{algorithmic}
\usepackage{graphicx}
\usepackage{caption}
\usepackage{subcaption}
\usepackage{textcomp}
\usepackage{xcolor}
\usepackage{multirow}
\usepackage[htt]{hyphenat}
\usepackage{hyperref}

\newcommand{\orcid}[1]{\href{https://orcid.org/#1}{\includegraphics[width=0.03\textwidth]{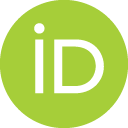}}}

\begin{document}
\title{Real-Time Multi-Object Tracking using YOLOv8 and SORT on a SoC FPGA}
\titlerunning{MOT with YOLOv8 and SORT}
%
\author{Michal Danilowicz \orcid{0000-0001-8851-8186} \and
Tomasz Kryjak \orcid{0000-0001-6798-4444}}
\authorrunning{M. Danilowicz, T. Kryjak}
%
\institute{Embedded Vision Systems Group, Computer Vision Laboratory, \\ Department of Automatic Control and Robotics, \\ AGH University of Krakow, Poland\\
\email{\{danilowi,tomasz.kryjak\}@agh.edu.pl}}

\maketitle

\begin{abstract}

Multi-object tracking (MOT) is one of the most important problems in computer vision and a key component of any vision-based perception system used in advanced autonomous mobile robotics.
Therefore, its implementation on low-power and real-time embedded platforms is highly desirable.
Modern MOT algorithms should be able to track objects of a given class (e.g. people or vehicles).
In addition, the number of objects to be tracked is not known in advance, and they may appear and disappear at any time, as well as be obscured.
For these reasons, the most popular and successful approaches have recently been based on the tracking paradigm. 
Therefore, the presence of a high quality object detector is essential, which in practice accounts for the vast majority of the computational and memory complexity of the whole MOT system.
In this paper, we propose an FPGA (Field-Programmable Gate Array) implementation of an embedded MOT system based on a quantized YOLOv8 detector and the SORT (Simple Online Realtime Tracker) tracker.
We use a modified version of the FINN framework to utilize external memory for model parameters and to support operations necessary required by YOLOv8.
We discuss the evaluation of detection and tracking performance using the COCO and MOT15 datasets, where we achieve $0.21$ mAP and $38.9$ MOTA respectively.
As the computational platform, we use an MPSoC system (Zynq UltraScale+ device from AMD/Xilinx) where the detector is deployed in reprogrammable logic and the tracking algorithm is implemented in the processor system.

\keywords{Multi-object tracking, MOT, Object detection, AI, SoC FPGA, FINN, YOLOv8}
\end{abstract}

\section{Introduction}

\begin{figure}
    \centering
    \includegraphics[width=1\linewidth]{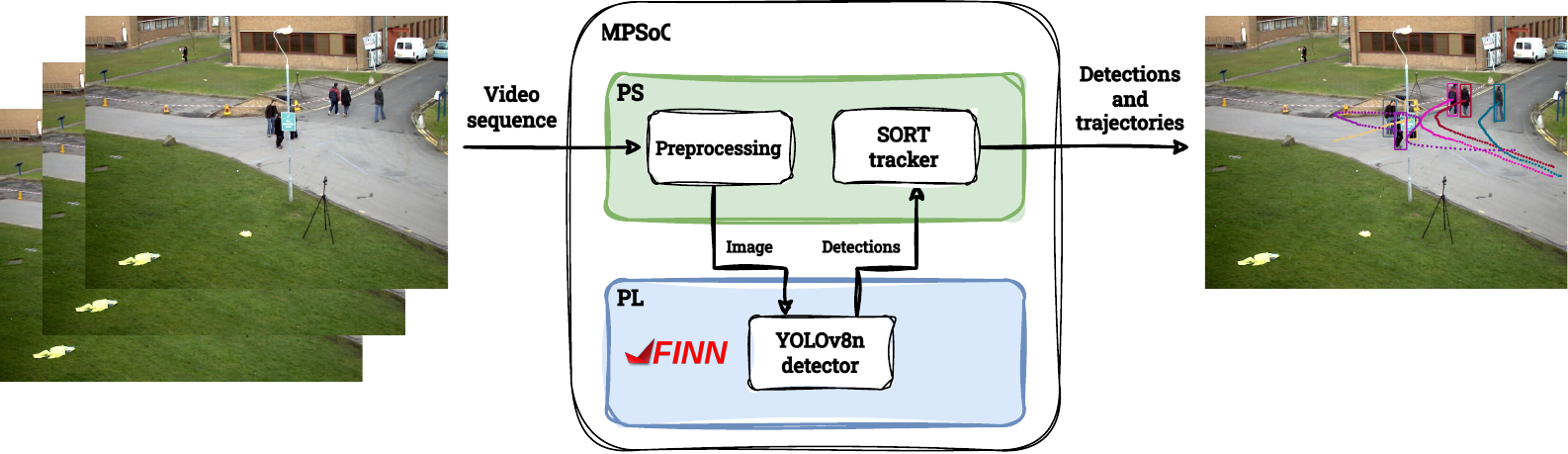}
    \caption{Top-level diagram of our multi-object tracking system implemented in SoC FPGA.}
    \label{fig:simple_toplevel}
\end{figure}



Multi-object tracking (MOT) is one of the most important issues in computer vision, due to its key role in the perception systems of Autonomous Vehicles (AV), Advanced Driver Assistance Systems (ADAS) or Advanced Video Surveillance Systems (AVSS). 
It allows to predict the trajectories of moving objects, such as pedestrians around a car.
Nowadays, most MOT methods are based on the tracking-by-detection paradigm: in the first step, the tracked objects are detected in the image, and in the second step, current detections are matched to the history of the previously tracked objects.
It should be noted that in the above scheme, the role of the detector is crucial -- it is the element on which the final effectiveness of the algorithm largely depends.
Moreover, this step it is usually also the most computationally and memory-intensive.

Currently, a deep neural network is usually used as the detection module.
On the one hand, this provides good results, but on the other hand, it requires a powerful computing platform -- e.g. a modern GPU (Graphic Processing Unit).
If the MOT system is to be applied to, for example, mobile robotics, it needs to run on a suitable embedded platform with a limited energy budget.
One of these are FPGA SoCs, for example Zynq UltraScale+ MPSoC (Multi-Processor System on Chip).

This paper presents the design and implementation of a hardware-software tracking system in an SoC FPGA device (Figure \ref{fig:simple_toplevel}).
A quantised YOLOv8 object detector is trained on the COCO \cite{coco} dataset to provide the system with precise detections.
It is implemented in the programmable logic (PL) using the FINN framework, which was adapted for implementation of the YOLOv8 architecture and to properly utilise external RAM for model parameters.
For the tracking part of the system, the SORT (Simple Online Real-time Tracking) algorithm computed on the processing system (PS) of the SoC is used. 

The main contribution of the paper can be summarised as allows:
\begin{itemize}
    \item We present an implementation of a real-time, embedded multi-object tracking-by-detection system with modular hardware and software architectures as a baseline for future research. We have achieved detection speed of $195.3$ fps.
    \item We emphasise the importance of a comparable evaluation of an embedded multi-object tracking system and evaluate our proposal on the commonly used and challenging MOT15 \cite{firstmot} and COCO datasets, where we achieve $38,9$ MOTA and $0.21$ mAP respectively.
\end{itemize}


The reminder of this papers is organised as follows. Section \ref{sec:previous_work} summarises previous work on multi-object tracking algorithms and their implementation in FPGAs. Section \ref{sec:algorithm} describes the proposed algorithm. Section \ref{sec:implementation} presents the proposed computational architecture on the SoC FPGA device. Section \ref{sec:evaluation} discusses the appropriate evaluation approach for embedded multi-object tracking systems and presents the results of detection and tracking quality of the proposed embedded tracker. Section \ref{sec:conclusion} summarises the work and indicates directions for future research.

\section{Previous work}
\label{sec:previous_work}

Multi-object tracking algorithms, due to their usefulness, are currently being intensively developed \cite{MOT_survey1}, \cite{MOT_survey2}.
As many solutions are currently being presented, there have also been many papers and challenges on datasets and metrics for MOT evaluation \cite{MOT_eval1}, \cite{MOT_eval2}, \cite{MOT_eval3}, \cite{MOT_eval4}.
What all solutions and challenges have in common, however, is that a multi-object tracking system needs as input a set of detections of objects of a specific class present in the scene.
In challenges based on the MOTChallenge sets \cite{firstmot}, \cite{MOT16}, \cite{MOT20}, the task is to track only pedestrians, while challenges using the BDD100k set \cite{BDD100k} require tracking and classification of a total of eight object classes (pedestrian, rider, car, bus, truck, train, motorbike, bicycle).
The detections in the MOT system are necessary to initialise the actual tracking algorithm, and many successful methods rely solely on the association of the input detections to the corresponding motion trajectories -- so-called tracking-by-detection approach \cite{MOT_survey_association}.

\begin{table*}[]
\centering
\caption{A summary of existing implementations of MOT systems in FPGAs.}
\label{tab:review}
\begin{tabular}{|c|c|c|c|c|}
\hline
Paper & \begin{tabular}[c]{@{}c@{}}Tracking\\ speed\end{tabular}                                                           & \begin{tabular}[c]{@{}c@{}}Detection\\ algorithm\end{tabular}                                                           & \begin{tabular}[c]{@{}c@{}}Tracking\\ algorithm\end{tabular}                                                                                                         & Evaluation                                                                                                                                        \\ \hline
\begin{tabular}[c]{@{}c@{}}2011\\ \cite{MOT_fpga62}\end{tabular} & --                                                                                                                 & \begin{tabular}[c]{@{}c@{}}not included\\ in the system,\\ targets initialized\\ by user\end{tabular}                     & \begin{tabular}[c]{@{}c@{}}three algorithms\\ that can be\\ reconfigured in runtime:\\ correlation tracker,\\ brightest spot tracker,\\  centroid tracker\end{tabular} & not reported                                                                                                                           \\ \hline
\begin{tabular}[c]{@{}c@{}}2013\\ \cite{MOT_fpga66}\end{tabular} & 60 fps                                                                                                             & \begin{tabular}[c]{@{}c@{}}moving object\\ detection by\\ background\\ modeling\\ and subtraction\end{tabular}            & particle filter per target                                                                                                                                           & \begin{tabular}[c]{@{}c@{}}qualitative results\\ on own sequences\end{tabular}                                                          \\ \hline
\begin{tabular}[c]{@{}c@{}}2015\\ \cite{MOT_fpga45}\end{tabular} & \begin{tabular}[c]{@{}c@{}}around\\ 10 - 3 fps\\ depending\\ on the\\ number of\\ objects\\ (10 - 50)\end{tabular} & \begin{tabular}[c]{@{}c@{}}not included\\ in the system,\\ using previously\\ generated regions\end{tabular}            & \begin{tabular}[c]{@{}c@{}}partition of bipartite\\ graph \\ representing targets\\ and detections,\\ based on region features\\ like position and shape\end{tabular}  & not reported                                                                                                                            \\ \hline
\begin{tabular}[c]{@{}c@{}}2015\\ \cite{MOT_fpga68}\end{tabular} & 1.3 fps                                                                                                            & \begin{tabular}[c]{@{}c@{}}moving object\\ detection by\\ background\\ modeling\\ and subtraction\end{tabular}            & Kalman filter per target                                                                                                                                             & \begin{tabular}[c]{@{}c@{}}qualitative results\\ on own sequences\end{tabular}                                                        \\ \hline
\begin{tabular}[c]{@{}c@{}}2017\\ \cite{MOT_fpga69}\end{tabular} & 152 fps                                                                                                            & \begin{tabular}[c]{@{}c@{}}color and shape\\ recognition\\ of predefined\\ markers\end{tabular}                           & \begin{tabular}[c]{@{}c@{}}targets are not\\ distinguishable\end{tabular}                                                                                                                                      & \begin{tabular}[c]{@{}c@{}}precision and recall rates\\ of 98\% on own sequences\end{tabular}                                          \\ \hline
\begin{tabular}[c]{@{}c@{}}2019\\ \cite{MOT_fpga17}\end{tabular} & 11.1 fps                                                                                                           & \begin{tabular}[c]{@{}c@{}}single-scale,\\ 5-anchor YOLO\\ with depthwise\\ separable\\ convolution\\ backbone\end{tabular} & \begin{tabular}[c]{@{}c@{}}a single detection with\\ highest confidence\\ score is considered\\ a tracking output\end{tabular}                                       & \begin{tabular}[c]{@{}c@{}}IoU metric on \\ a single object \\ detection dataset\\ DAC SDC \cite{dac_sdc}\end{tabular}                                 \\ \hline
\begin{tabular}[c]{@{}c@{}}2021\\ \cite{MOT_fpga70}\end{tabular} & 36.2 fps                                                                                                           & \begin{tabular}[c]{@{}c@{}}SSD detector every\\ N frames,\\ trained on the\\ VOC dataset\end{tabular}                         & \begin{tabular}[c]{@{}c@{}}per-target KCF tracker\\ between detection frames\end{tabular}                                                                            & \begin{tabular}[c]{@{}c@{}}single object tracking\\ performance\\  on OTB-10 dataset and \\ detection performance\\ on VOC\end{tabular}      \\ \hline
\begin{tabular}[c]{@{}c@{}}2023\\ \cite{MOT_fpga84}\end{tabular} & 91.65 fps                                                                                                           & \begin{tabular}[c]{@{}c@{}}2-scale\\ YOLOv3-tiny \\ detector\end{tabular}                         & \begin{tabular}[c]{@{}c@{}}Deepsort \cite{deepsort}\end{tabular}                                                                            & \begin{tabular}[c]{@{}c@{}}detection and MOT\\ performance \\ evaluated on choosen sequences \\ using mAP and CLEARMOT\\ metrics from the\\ ua-detrac \cite{ua_detrac} dataset. \end{tabular}   \\ \hline
\begin{tabular}[c]{@{}c@{}}2023\\ \cite{MOT_fpga88}\end{tabular} & 45 fps                                                                                                           & \begin{tabular}[c]{@{}c@{}}YOLOv4 detector\end{tabular}                         & \begin{tabular}[c]{@{}c@{}}Deepsort \cite{deepsort}\end{tabular}                                                                            & \begin{tabular}[c]{@{}c@{}}detection and MOT \\ evaluation not reported \end{tabular}   \\ \hline
\begin{tabular}[c]{@{}c@{}}2023\\ \cite{MOT_fpga90}\end{tabular} & 30 fps                                                                                                           & \begin{tabular}[c]{@{}c@{}}not included\end{tabular}                         & \begin{tabular}[c]{@{}c@{}}auction-based \\ assignment algorithm\end{tabular}                                                                            & \begin{tabular}[c]{@{}c@{}} MOT performance evaluated \\ on the MOT15 dataset \end{tabular}  \\ \hline
\textbf{Ours} & \begin{tabular}[c]{@{}c@{}} 195.3 fps \\ for \\ detection, \\ 24 fps \\ for whole \\ system \end{tabular}                                                                                                           & \begin{tabular}[c]{@{}c@{}}YOLOv8n\\ detector \\ \end{tabular}                         & \begin{tabular}[c]{@{}c@{}}SORT \cite{deepsort}\end{tabular}                                                                            & \begin{tabular}[c]{@{}c@{}}detection and MOT\\ performance \\ evaluated on COCO, ua-detrac \\ and MOT15 datasets \\ using mAP and CLEARMOT \\ metrics\end{tabular}     \\ \hline
\end{tabular}
\end{table*}

\subsubsection{Hardware implementations}

Existing work on multi-object tracking on FPGA platforms often lacks a high-quality detector capable of indicating the presence and position of objects of a specific class.
In papers \cite{MOT_fpga45}, \cite{MOT_fpga90} and \cite{MOT_fpga62}, due to the absence of a detector, the systems cannot autonomously initialise the tracking algorithm, so they can only be used as some part of a larger perception system.
In the work \cite{MOT_fpga69}, only objects on which a special marker has been placed can be tracked.
In the papers \cite{MOT_fpga66} and \cite{MOT_fpga68}, background modelling and subtraction algorithms are used for detection (so-called foreground object segmentation and detection).
This method of detection, however, excludes its use in autonomous robots/vehicles, which require perception in a dynamically changing environment and in moving camera scenario.
In addition, it is characterised by losing objects that stop moving, high detection noise and, by themselves, do not offer any object classification.

Some of the existing works, lack adequate evaluation of tracking and detection quality using standardised datasets and metrics, making it impossible to effectively compare a given solution with existing ones.
FPGA implementations require certain trade-offs, like fixed-point computations or the use of a more compact neural network model, in the tracking algorithm and it is necessary to evaluate their impact on tracking quality.
Without proper evaluation, it is difficult to draw conclusions about what platform, computing power and power consumption are needed to achieve the desired quality of an embedded multi-object tracking system.
A further discussion on evaluation of embedded MOT systems is presented in Section \ref{sec:evaluation}.

The papers \cite{MOT_fpga17} and \cite{MOT_fpga70} contain a high quality detector, however, the effectiveness of the first system is only evaluated in terms of single object detection quality on the DAC SDC \cite{dac_sdc} set, while the second system is evaluated on the single object tracking set.
It is therefore impossible to compare with these publications in terms of performance on a multi-object tracking task.
In the paper \cite{MOT_fpga90}, the authors do not present any evaluation of detection or tracking quality.
They present a hardware implementation of the YOLOv4 detector and a similarity matrix calculation system for the Deepsort \cite{deepsort} algorithm.
The paper, however, points out that there is no processor system to handle the hardware acceleration.

In summary, excluding solutions that are not complete MOT tracking systems, do not have a suitable detector and those that cannot be directly compared with due to the lack of MOT evaluation, only the work \cite{MOT_fpga84} should be considered a direct competitor of our proposal.
The authors present an embedded MOT system based on the Deepsort algorithm, which uses a YOLOv3-tiny 2-scale detector.
The processing speed allows real-time perception at $91.65$ fps, and its multi-object tracking performance is demonstrated on three selected sequences from the UA-DETRAC \cite{ua_detrac} set representing traffic.
In Section \ref{sec:evaluation}, we will present a comparison of our solution with the above.
A review of existing solutions is summarised in Table \ref{tab:review}.

\section{The proposed MOT algorithm}
\label{sec:algorithm}


As a baseline for our tracking system we have adapted the original implementation of the SORT (Simple Online Real-time Tracking) algorithm which was first presented in the paper \cite{sort}.
The method utilises the Kalman filter, in which the prediction of the state of each tracked object obtained from a linear motion model is corrected by a measurement of the object's position obtained from the detector.
The challenge is then to appropriately assign new detections in each image frame to the currently tracked objects.
One of the simplest ways to solve this problem is to use the Hungarian algorithm, using the IoU (Intersection over Union) metric between detections and objects as the assignment weights.
SORT is the basis for many of the more advanced methods (DeepSORT \cite{deepsort}, BoT-SORT \cite{bot_sort}, ByteTrack \cite{bytetrack}), and in itself offers very good tracking results at a relatively low computational cost.

\subsection{Detector}



\begin{figure}
    \centering
    \includegraphics[width=1\linewidth]{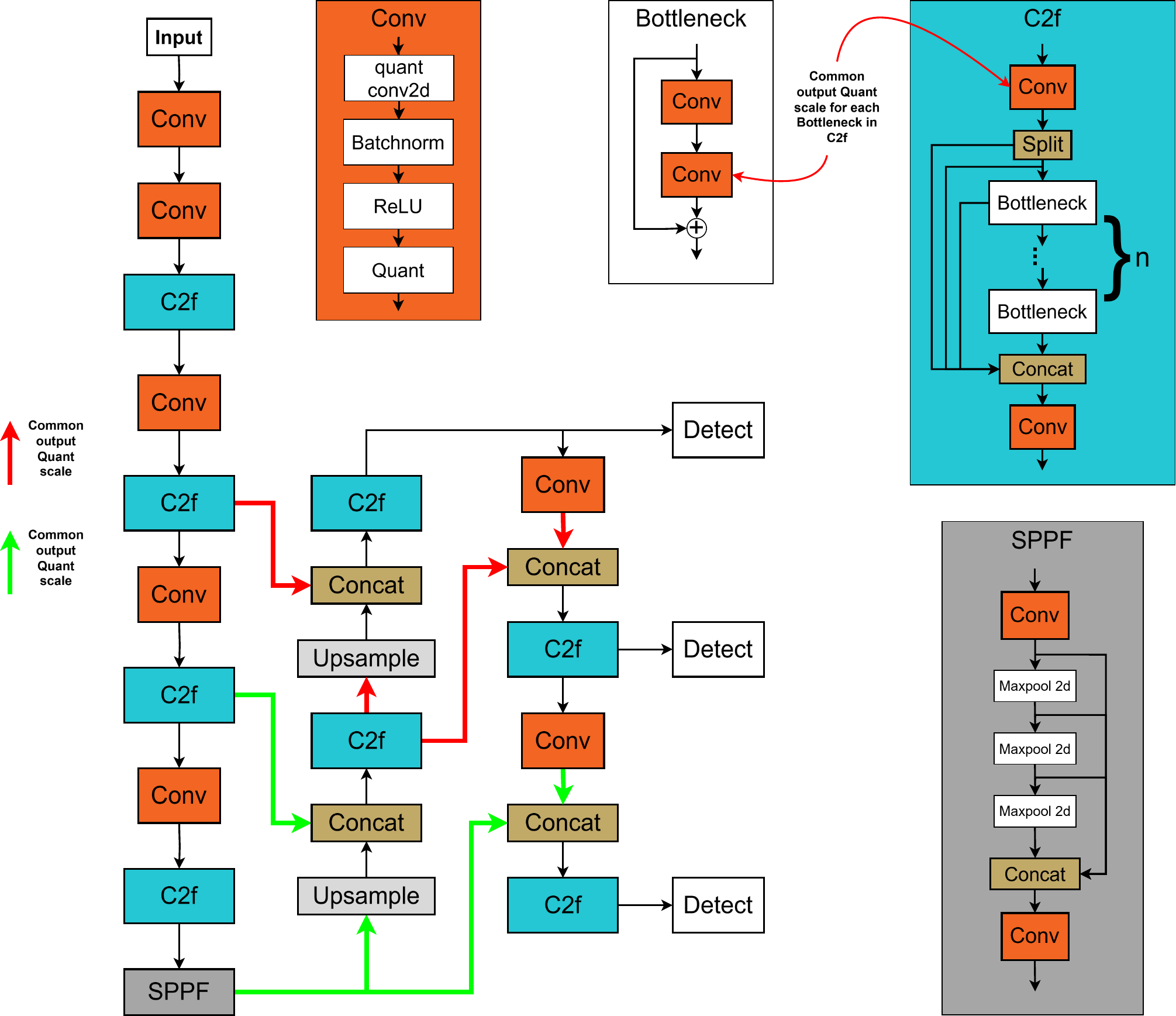}
    \caption{Topology of the YOLOv8 detector used in the proposed MOT system. The \texttt{Conv} block contains a quantised 2d convolution, batch normalisation, ReLU and activation quantiser. Red arrows and green arrows represent two groups of quantised tensors, that, each share a common quantisation scale. This was necessary to properly simplify the computational graph -- more details this in Section \ref{sec:fpga_arch}}
    \label{fig:yolov8_top}
\end{figure}

The YOLOv8\_n (nano) detector ($3.2$M parameters), which is one of the latest detectors in the YOLO (You Only Look Once) family, was used in this study.
It is a representative of the one-step detectors ``family'', in which the convolutional features of the whole image serve as input to a box predictor, and the class membership classification and regression of surrounding rectangles is performed for each pixel.
Unlike previous versions of detectors of this kind, YOLOv8 is an anchor-free detector, i.e. it does not require a predefined set of bounding boxes.
Instead, a probability distribution is predicted for each of the four parameters of the bounding box, the expected value of which is the result of the detection of \cite{dfl_head}.
Another difference introduced in YOLOv8 is the use of a single joint representation score instead of separately predicting the probability of class membership and objectness score.
The detector architecture is shown in Figure \ref{fig:yolov8_top}.
After adapting the original implementation \cite{ultralytics_github} for quantised training, the model was trained on the COCO using the procedure discussed in Section \ref{sec:quantization}.




\subsection{Quantization}
\label{sec:quantization}


The implementation of such deep models in full 32bit floating-point precision may not be possible on embedded platforms due to limited computational and memory resources.
For this reason, quantisation of parameters and intermediate results (activations) to an integer representation with fewer bits is usually used.
A smaller number of bits saves resources for storing model parameters (or reading from external memory) and for multiplication and accumulation operations.
In addition, operations performed on integers are much less complex than operations on floating point numbers.
However, these savings usually come at the expense of accuracy \cite{quantization,quantization2}.

The simplest approach is post-training quantisation, in which the parameters of a full-precision learned model are quantised to an appropriate representation before deployment.
This approach usually gives worse results than including quantisation during training (so-called Quantisation-Aware Training, QAT) \cite{quantization2}.
With QAT training, the model parameters are stored at full precision, however, during forward propagation, they are quantised with the input of each layer to the appropriate numerical representation before performing multiplication and accumulation operations.
In this way, integer operations of the target platform are simulated.
During back-propagation, full-precision weights are modified based on the gradients of the quantised operations.

The applied in this work QAT implementation uses the \texttt{Brevitas} library, which is a wrapper for the popular PyTorch library and supports the operations discussed above.
It is important to remember to quantise the intermediate results after each convolution layer, as the operations of multiplication and accumulation of integers increases the number of bits for their representation.

First, the model was trained to full precision on the COCO set until convergence.
An SGD (Stochastic Gradient Descent) optimiser was used with an initial learning rate of $0.01$, which decreased to $0.0001$ for 300 epochs.
The input image was scaled to $320$ pixels for the longer edge.
Then, starting with the weights at full precision, the model was fine-tuned with quantisation enabled to 4 bits per weight and activation.
During fine-tuning, a constant learning rate of $0.0001$ was applied and the EMA (Exponential Moving Average) of all model parameters was tracked.
For evaluation, the model was loaded with these averaged parameters.


\section{Hardware implementation}
\label{sec:implementation}



\begin{figure}
    \centering
    \includegraphics[width=0.9\linewidth]{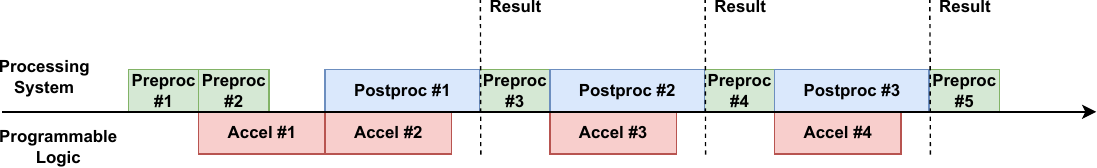}
    \caption{Parallel task sharing between PS and PL in the SoC FPGA device. The length of tasks here is not precise, it only illustrates that the PL part is faster than the PS part in our case.}
    \label{fig:parallel_PS_PL}
\end{figure}

\begin{figure}
    \centering
    \includegraphics[width=0.5\linewidth]{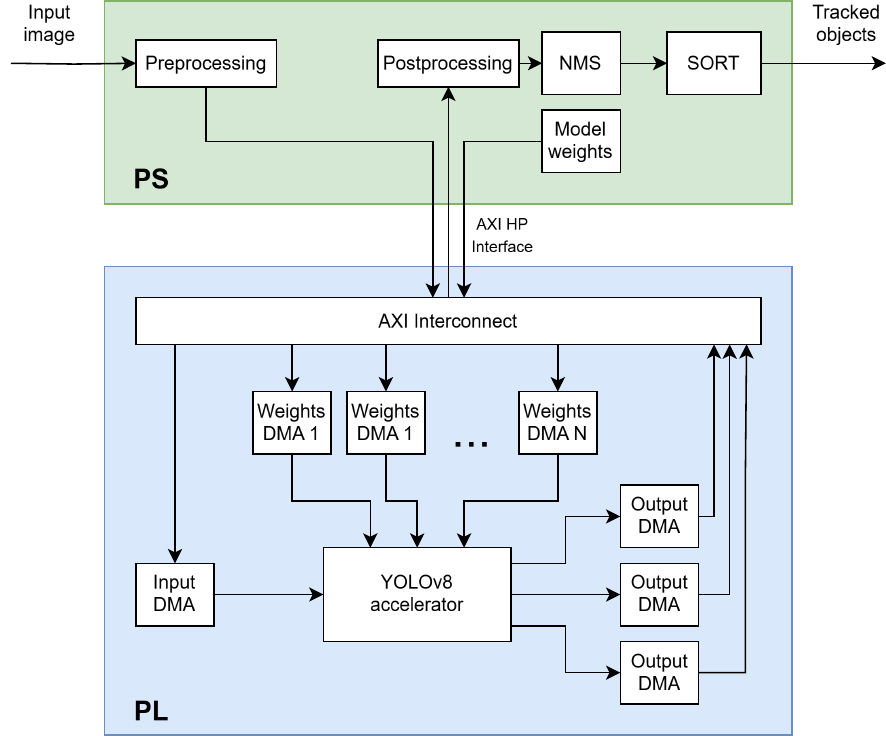}
    \caption{Hardware-software system implemented in the ZCU102 SoC FPGA.}
    \label{fig:top_level}
\end{figure}

Since in the adopted multi-object tracking algorithm, the detection step is performed independently of the tracking result, it is possible to separate these tasks into the programmable logic (PL) and the processor system (PS) of the FPGA SoC device (Figure \ref{fig:parallel_PS_PL}).
Preprocessing involves scaling the input image to the appropriate size and sending it via DMA to the accelerator in the PL.
During postprocessing, the three feature maps obtained from the three detection heads are interpreted into boudning boxes, which are subjected to NMS (Non-Maximal Suppression) filtering.
The SORT algorithm step is also regarded as a postprocessing step.
The top-level diagram of the entire system is shown in Figure \ref{fig:top_level}.


\subsection{FPGA architecture}
\label{sec:fpga_arch}

The FINN framework \cite{FINN} was used to implement the detector architecture in the FPGA logic of the SoC FPGA device.
It is a framework that includes Python libraries such as \emph{finn-hlslib}, \emph{finn-rtllib}, \emph{brevitas} and \emph{qonnx}.
With FINN, it is possible to conveniently perform operations on \emph{qonnx} graph nodes such as moving some operation past another and folding repeated neighbouring operations into one.
FINN also provides templates to call the appropriate \emph{finn-hlslib} library function (or RTL module from the \emph{finn-rtllib library}) for each node in the graph representation of the \emph{onnx} model.

\begin{figure}
    \centering
    \includegraphics[width=0.5\linewidth]{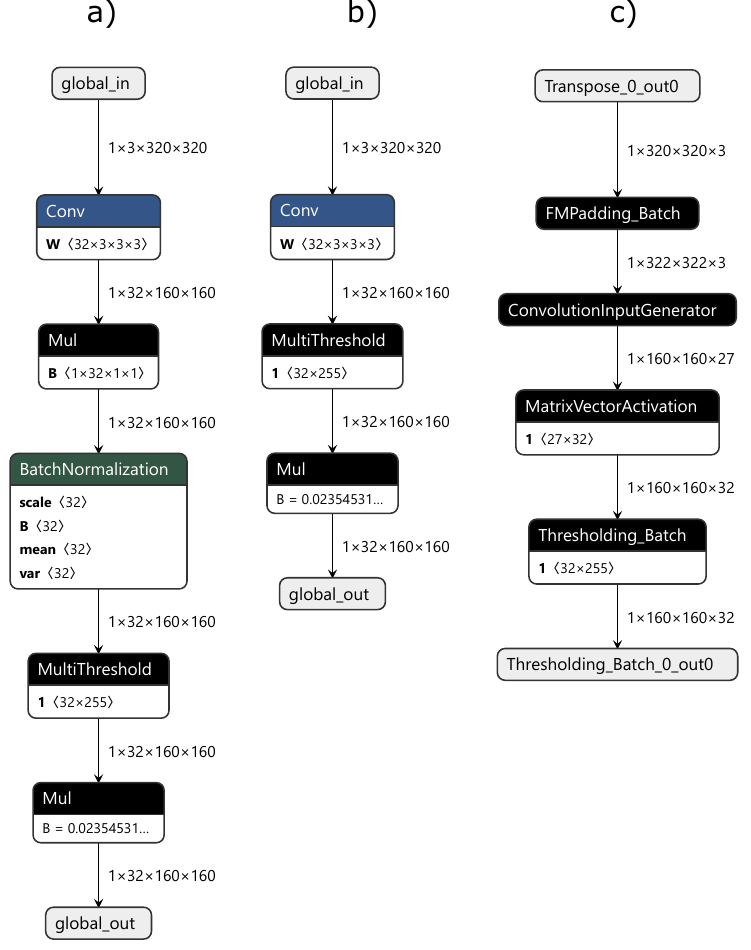}
    \caption{Deployment of a single \texttt{conv} block to an FPGA using the FINN library. First, the block is loaded from \emph{brevitas} code to graph representation \textbf{a)}. Then, affine transformations are collapsed into the MultiThreshold operation and the Mul node at the end can be moved past the following convolution in the network or delegated to postprocessing on the PS if this is the end of the accelerator \textbf{b)}. Finally, each node is represented by IP from \emph{finn-hlslib} library (convolution is a sequence of padding, context generation and matrix multiplication) \textbf{c)}.}
    \label{fig:convbnrelu_finn}
\end{figure}

The following steps of the procedure for implementing the \texttt{conv} block used in YOLOv8 are shown in Figure \ref{fig:convbnrelu_finn}.
Graph \ref{fig:convbnrelu_finn}a represents a block imported directly from the Python code used for network training.
The multiplication operation \texttt{Mul} results from quantisation and transforms an integer into a real-valued representation.
The scaling factor of \texttt{Mul} depends on the adopted range of real numbers to be quantised and the number of quantisation thresholds ($2^N - 1$ for the uniform $N$ bit quantisation used here).
The \texttt{MultiThreshold} block implements the requantisation.
The module performs three functions: it implements a quantiser; an activation function; and it allows to implicitly compute the preceding affine operation  \cite{streamline}.
The input of the module is compared against a set of thresholds $T = \{t_0, t_1, .... t_n\}$ and the output is the index of the smallest $t_i$ that is bigger than the input.
The calculation of an affine operation $ax + b$ can be omitted and implicitly performed by the \texttt{MultiThreshold} by substituting $t_i \gets (t_i - b)/a$.

\begin{figure}
    \centering
    \includegraphics[width=0.6\linewidth]{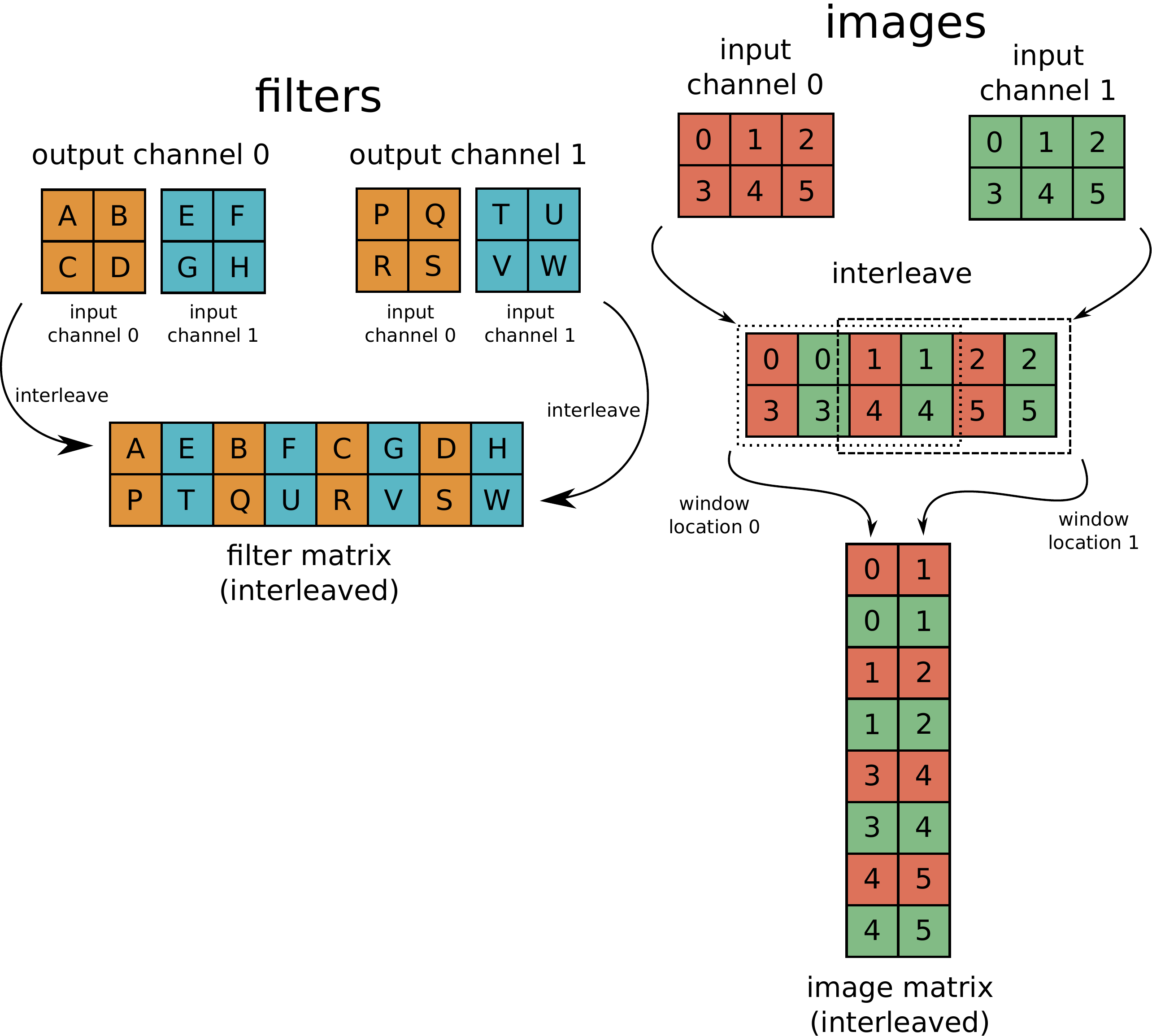}
    \caption{The diagram shows the computation of a standard convolution of $2 \times 2$ with two input and output channels for an input of $2 \times 3$. Each row of the filter matrix corresponds to the weights needed to compute one output channel. In contrast, each column of the image matrix is composed of the input data located in one position of the filter context \cite{jlpea_nasze}.}
    \label{fig:convolution}
\end{figure}

The \texttt{conv} block is decomposed into padding, context generation and matrix-vector multiplication operations (Respectively, \texttt{FMPadding\_Batch}, \texttt{ConvolutionInputGenerator} and \texttt{MatrixVectorActivation} blocks in Figure \ref{fig:convbnrelu_finn}c).
The operation performed by the convolution layers is in practice a matrix multiplication operation (an example for a 2x2 convolution for a 2x3 input is shown in Figure \ref{fig:convolution}).
The processed channels are interleaved, thus avoiding the caching of scalar products from individual input channels to be later summed to obtain a single output channel.

The balance between the degree of parallelization and the accelerator's resource consumption can be controlled by selecting the number of input channels processed in parallel (SIMD -- Single Instruction Multiple Data) and the number of computational elements (PE).
The rows of the filter matrix (Figure \ref{fig:convolution}) are distributed in parallel between the PEs (hence the number of output channels must be a multiple of the PEs), while the columns are distributed between the SIMD input lines (hence the number of input channels is a multiple of the SIMD).

The affine operations resulting from batch normalisation and quantisation can be directly moved past the convolution operation due to its linear nature.
In blocks with skip-connections, like \texttt{C2f}, affine operations can be shifted behind the fork node by copying the node to the beginning of each branch. 
In the case of a join node behind a bottleneck block, the operations must be identical before being moved past the join.
This condition was met by restricting the quantization scales of some activations to be the same during training.
To be specific, in each C2f block, the scale of the quantised output of the first \texttt{Conv} block and the second \texttt{Conv} inside each \texttt{Bottleneck} need to be the same.
For the same reason, some tensors at the highest level of the YOLOv8 topology also share common quantisation scale -- denoted by the same colour in Figure \ref{fig:yolov8_top}.
These affine operations are moved down the computational graph until all have been removed from the graph by joining them with \texttt{MultiThreshold} nodes using method described earlier.



To enable acceleration of the YOLOv8 detector in the FINN framework, the library has been enhanced with the channel split operation \texttt{Split}, and the channel concatenation operation \texttt{Concat} has been updated to accept inputs with different data types.
This situation occurs in the \texttt{C2f} block, where 4-bit activations from the \texttt{Split} block need to be concatenated with 5-bit activations from the \texttt{Bottleneck} block.


One of the challenges of the pipelined accelerator architecture used here is that certain adjacent operations in the pipeline -- even though they have the same throughput -- may not work fully efficiently.
This is due to the fact that for certain operations (matrix multiplication, for example), the ``producer'' accumulates input data for a while and then wants to send a lot of output data in one burst.
This burst may be too large to be read quickly enough by the ``consumer'', causing the pipeline to stop temporarily.
To avoid this, a FIFO queue deep enough to accommodate the burst in question should be placed between the ``producer'' and the ``consumer''.


Another problem is the need to balance the parallel branches of the graph also by placing FIFOs with appropriate depths.
This avoids deadlocks, in which one branch halts all processing before a stream addition operation because the data from the other branch is not yet ready (because the stream is halted).
The appropriate FIFO depths needed for smooth accelerator operation are selected by placing deep FIFOs in front of each branch of the graph and then simulating to check the largest saturation of each FIFO.
Our detector in the reprogrammable logic uses one static configuration for input image shape of $320 \times 192$ pixels.

\subsection{Processing System}




The processor part of the system receives data from the detector in the form of three feature maps - one for each detection head (Figure \ref{fig:yolov8_top}).
Each pixel location on those feature maps contains $84$ channels - $4$ for the regression of the bounding box, and the classification distribution for all $80$ classes from the COCO dataset.
The resulting detections are thresholded by the probability score and then subjected to non-maximum suppression filtering.
The detections of the selected object classes are passed to the SORT algorithm.
As the AMD PYNQ operating system was used on the SoC FPGA platform, it was possible to program the processor system in the Python language and using popular implementations of the Kalman filter and the Hungarian algorithm from the \texttt{filterpy} and \texttt{scipy} libraries.

It is also the task of the processor system to handle the DMA modules appropriately, as a dedicated DMA is assigned for each \texttt{MatrixVectorActivation} module to send convolutional layer weights.
All filter parameters need to be sent by the DMA for each convolutional filter position on the input feature tensor.


\section{Evaluation}
\label{sec:evaluation}



For the evaluation of state-of-the-art MOT methods, the most commonly used datasets are MOT15, MOT16, MOT17 and MOT20 \cite{MOT_survey1} using HOTA \cite{MOT_eval2} and CLEARMOT \cite{MOT_eval3} metrics.
To isolate the impact of detector quality on tracking performance during MOT evaluation, the set of detections in each frame of the sequence is fixed for all algorithms tested.
This is an evaluation using so-called public detections and, thanks to this approach, multi-object tracking studies can be focused on the association algorithm itself.

However, in practical MOT applications, it is necessary to take into account the presence of a detector in the system -- especially in embedded systems, as the detection algorithm is responsible for the vast majority of the computational complexity of the entire system.
It is also worth noting that the tracking algorithm is only able to predict the trajectories of those objects that the detector is able to detect -- so-called private detections.
For these reasons, in order to correctly infer the effectiveness of a given embedded MOT system solution against others, it is necessary to evaluate MOT on popular datasets from the MOTChallenge using private detections.



In addition to MOT evaluation using private detections, the evaluation of the stand-alone detector used in the system is also needed for a comprehensive analysis of the embedded solution.
For some specific application of the embedded MOT system, it may be helpful to additionally train the detector on a set containing sequences representing objects under the conditions in which the system will ultimately operate \cite{MOT_fpga66}, \cite{MOT_fpga68}, \cite{MOT_fpga84}.
However, reporting results on such datasets does not allow for an effective comparison of the system among state-of-the-art.
In order to be able to absolutely assess the ability of a given system to detect the objects to be tracked, popular datasets such as MS COCO \cite{coco} should be used for evaluation and training.
For evaluating detection quality, we use the primary metric in the COCO challenge - mAP (mean average precision).
It describes detection precision averaged over all object categories and 10 IoU thresholds between $0.5$ and $0.95$.

The detector architecture shown in Figure \ref{fig:yolov8_top} was trained on COCOtrain and evaluated on COCOval, the results are summarised in Table \ref{tab:coco_eval}.


\begin{table}[]
\centering
\caption{Detection evaluation of our quantized YOLOv8 detection on the COCOval dataset. 4w4a denotes 4 bits for weights and 4 bits for activations.}
\begin{tabular}{|c|c|c|c|c|}
\hline
Model & \begin{tabular}[c]{@{}c@{}}Our\\ full precision\end{tabular} & \begin{tabular}[c]{@{}c@{}}Our\\ 4w4a\end{tabular} & \begin{tabular}[c]{@{}c@{}} Faster-RCNN \cite{faster_rcnn} \\ full precision \end{tabular}               & \begin{tabular}[c]{@{}c@{}}2018 \cite{ssdlitem2_fpga}\\ 8w8a\end{tabular} \\ \hline
Hardware & CPU & FPGA & CPU & FPGA \\
\hline
\begin{tabular}[c]{@{}c@{}}COCOval\\ mAP\end{tabular} & 0.28                                                         & 0.21                                               & \multicolumn{1}{c|}{0.21} & 0.20                                                                 \\ \hline
\begin{tabular}[c]{@{}c@{}}Model size\end{tabular} & 96.4Mb                                                         & 12.0Mb                                               & \multicolumn{1}{c|}{1404Mb} & 25.5Mb                                                                 \\ \hline
\end{tabular}
\label{tab:coco_eval}
\end{table}

The well-known Faster-RCNN \cite{faster_rcnn} two-stage detector and a sequential model of similar complexity implemented in FPGA in \cite{ssdlitem2_fpga} are also listed for comparison.
Faster-RCNN is used for generation of public detections used to evaluate state-of-the-art MOT algorithms in MOT Challenge.
Note that the proposed detector achieves an mAP value of 0.21, which is at the same level as the more computational complex Faster-RCNN using floating-point computations and surpassed the result of \cite{ssdlitem2_fpga} with 8-bit quantisation for weights and activations.
However, with respect to the original floating-point model the drop in mAP is substantial -- around 25\%.

In competitions, the MOTA metric described by the Equation (\ref{eq:mota}), is used.
\begin{equation}
    MOTA = 1 - \frac{\sum_{t} (FN_t + FP_t + IDSW_t)}{\sum_{t} g_t}
    \label{eq:mota}
\end{equation}
It penalises the tracker for missed detections ($FN_t$), incorrect detections ($FP_t$) and a number of identity switches of tracked objects ($IDSW_t$) in each frame $t$.
The number of ground-truth objects in frame $t$ is denoted by $g_t$.
For the implemented 4-bit detector, an evaluation was performed on MOT15 set using CLEARMOT metrics.
Although there are no FPGA implementations of MOT systems evaluated with such methodology, we can compare our results with the original, non-embedded implementation of the SORT algorithm with Faster-RCNN detector evaluated on the train split of the MOT15 dataset in \cite{sort}.
We achieved combined MOTA of $0.389$, while the baseline SORT is reported to reach $0.340$.
Existing state-of-the-art algorithms offer higher quality tracking; however, their computational complexity does not allow application to embedded real-time systems.


In order to compare with the work \cite{MOT_fpga84}, an evaluation was performed on the same three sequences from the ua-detrac set.
The results are presented in Table \ref{tab:sota_comparison}.
We achieve comparable multi-object tracking quality using a significantly simpler MOT algorithm.
The FPGA detection processing speed of our detector was $195.3$fps for $300$MHz FPGA clock.
The performance was measured by clock cycles counting.
The running time of the SORT and NMS algorithm on the PS ZCU102 limits the speed of the overall system to around $24$ fps.
However, there is still room for improvement, as the operations are currently implemented in Python.



\begin{table}[]
\centering
\caption{Comparison of tracking performance for the sequences using in the paper \cite{MOT_fpga84}. MOTA metrics are presented.}
\begin{tabular}{|l|l|l|l|}
\hline
\multicolumn{1}{|c|}{\begin{tabular}[c]{@{}c@{}}sequence\\ name\end{tabular}} & MVI\_40701 & MVI\_40701 & MVI\_40701 \\ \hline
Ours                                                                          & 0.666      & 0.756      & 0.337      \\ \hline
\cite{MOT_fpga84}                                                                        & 0.675      & 0.696      & 0.405      \\ \hline
\end{tabular}
\label{tab:sota_comparison}
\end{table}



\begin{figure}
     \centering
     \begin{subfigure}[b]{0.24\textwidth}
         \centering
         \includegraphics[width=\linewidth]{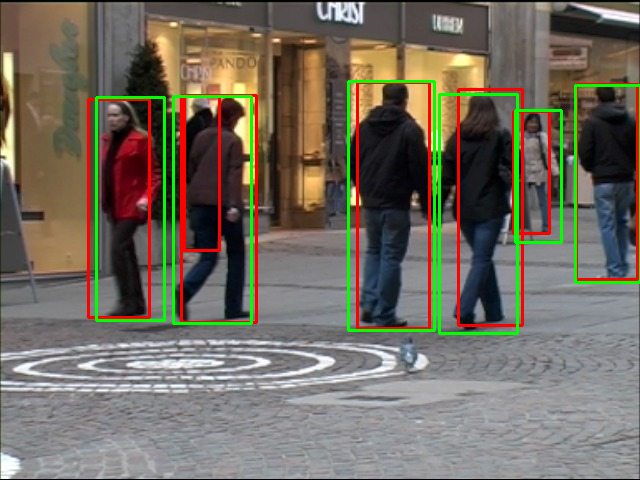}
     \end{subfigure}
    \begin{subfigure}[b]{0.24\textwidth}
        \centering
        \includegraphics[width=\linewidth]{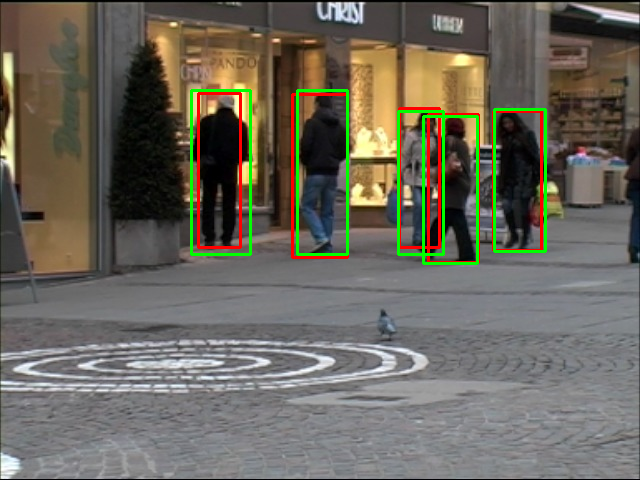}
    \end{subfigure}
     \begin{subfigure}[b]{0.24\textwidth}
         \centering
         \includegraphics[width=\linewidth]{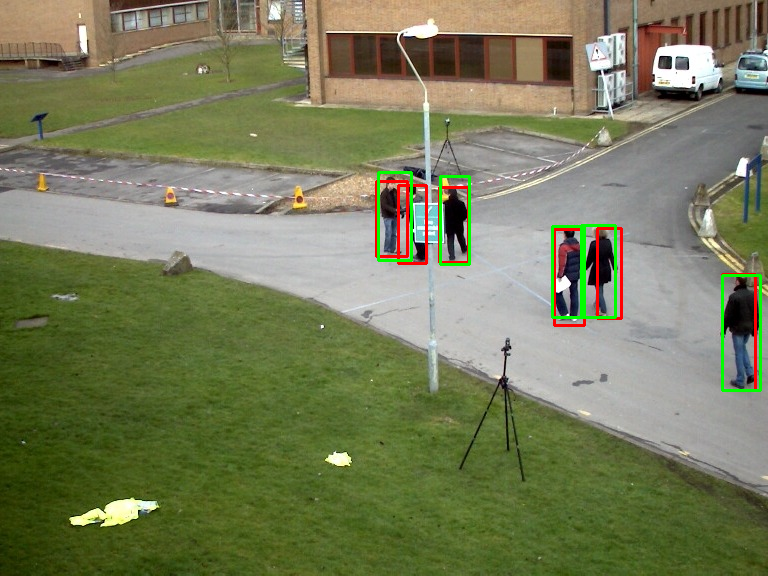}
     \end{subfigure}
    \begin{subfigure}[b]{0.24\textwidth}
        \centering
        \includegraphics[width=\linewidth]{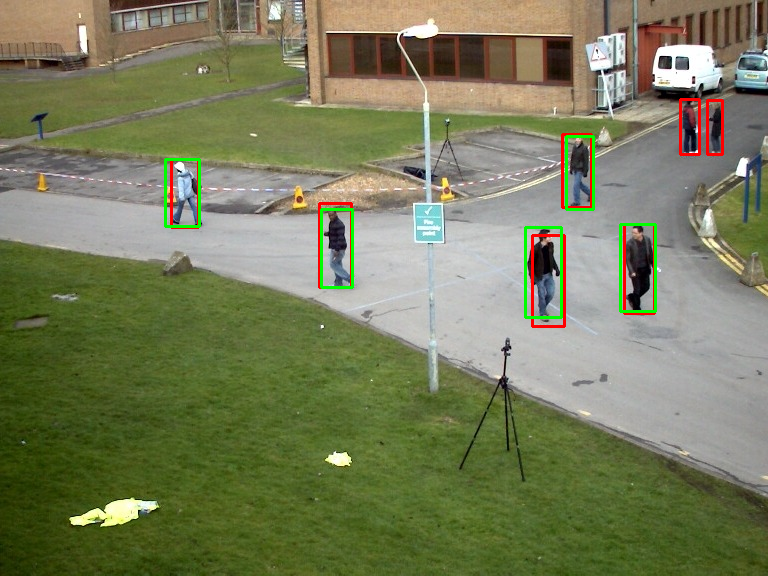}
    \end{subfigure}
        \caption{Visualized tracking result for TUD-Stadtmitte (left) and PETS09-S2L1 (right) sequences.
        MOT15 groundtruth is denoted by red and tracking output by green.}
        \label{fig:visualised2}
\end{figure}

Qualitative results of our system for example MOT15 sequences are shown in Figure \ref{fig:visualised2}.


The resource consumption of the implemented detector for 4-bit quantisation is shown in Table \ref{tab:utilisation}.
The advantage of pipelined fine-grained architectures is the high processing speed due to the full parallelization of all operations during video stream processing.

\begin{table}[]
\centering
\caption{FPGA resources utilisation for the whole design with axi\_interconnects and DMA modules and just for the detector. Absolute and relative to all available resources on the ZCU102 developement board. We also include utilisation reported in \cite{MOT_fpga84}, where the relative utilisation is for Zynq-700 device.}
\begin{tabular}{|l|l|l|l|l|}
\hline
                                                               & LUT     & FF      & BRAM    & DSPs    \\ \hline
Full design                                                   &205k (74.71\%) & 246k (44.94\%) &421 (46.16\%) &486 (19.29\%) \\ \hline
Detector only &105k (38.24\%) & 72k (13.08\%) &293 (32.13\%) &482 (19.13\%) \\ \hline
Work \cite{MOT_fpga84}                                                   &38k (22.2\%) & 43k (12.5\%) &132.5 (26.5\%) &144 (16.2\%) \\ \hline
\end{tabular}
\label{tab:utilisation}
\end{table}




Disadvantages, however, include the high computing resource requirements for the DMA modules and the interconnect modules connecting them to the PS-PL AXI4 bus of the SoC.
This applies in the case of using the RAM of the processor system to store the weights of the detector in reprogrammable logic as in this work.
Due to the need to place FIFO queues in the pipeline (for reasons described in Section \ref{sec:fpga_arch} and the need to requantise the activation after each convolution, the depth of the model also translates into a direct increase in memory resource requirements (BRAM, LUTRAM).


\section{Conclusion}
\label{sec:conclusion}


This paper presents a multi-object tracking system implemented on a heterogeneous SoC FPGA device.
The FINN library was used to implement the 4-bit quantised YOLOv8 nano detector architecture in reprogrammable logic, which was adapted to support external memory for storing the model parameters.
Thanks to FINN's modular fine-grained accelerator architecture, the resources allocated to individual operations can be selected.
For example, the way the parameters are stored (BRAM, URAM -- only for the Zynq UltraScale+ devices, LUTRAM or external memory) for the \texttt{MultiThreshold}, \texttt{MatrixVectorActivation} modules or the way multiplication and accumulation (LUT, DSP) are implemented for the \texttt{MatrixVectorActivation} module.
The degree of computational parallelization can also be selected depending on the hardware platform chosen.
To save memory resources, the requantisation operation \texttt{MultiThreshold} (which accounts for around $35$\% BRAM utilisation) could be implemented by multiplication using DSP resources which are currently underutilised (Table \ref{tab:utilisation}).
This will be part of the future work which would enable targeting smaller devices.



In addition to external memory management, the SORT algorithm was implemented in the processor system, which forms the basis of multi-object tracking algorithms using tracking-by-detection paradigm.
We achieve state-of-the art detection and tracking quality at high processing speed of the detector accelerator - $195.3$ fps.
This speed overhead could be utilised for processing bigger images for even higher detection quality.
Python implementation of the software being run on the processing system will be optimised in the future work.

The tracking algorithm in the processor system can be extended with more sophisticated association mechanisms based on the appearance features of the objects \cite{deepsort} or use correlation filters \cite{nasz_cf} to predict the trajectory if the detection of the tracked object fails.
The additional computation resulting from the network computing embeddings or correlation filter responses can then also be accelerated in reprogrammable logic.


\bibliographystyle{splncs04}
\bibliography{mybibliography}

\end{document}